\documentclass[runningheads]{llncs}
\newif\ifarxiv
\arxivtrue

\ifarxiv
\usepackage{eccv}
\else
\usepackage[review,year=2026,ID=12270]{eccv}
\fi

\usepackage{eccvabbrv}
\PassOptionsToPackage{hyphens}{url}
\usepackage{graphicx}
\usepackage{booktabs}
\usepackage{soul}
\usepackage[table]{xcolor}
\usepackage{makecell}
\usepackage{pifont}
\usepackage{enumitem}
\usepackage{algorithm}
\usepackage{algpseudocode}
\IfFileExists{axessibility.sty}{\usepackage[accsupp]{axessibility}}{}

\ifarxiv
\usepackage[breaklinks,hidelinks]{hyperref}
\else
\usepackage[pagebackref,breaklinks,colorlinks,citecolor=eccvblue]{hyperref}
\fi

\usepackage{multirow}

\newcommand{\cmark}[1]{\textcolor{green!60!black}{\textbf{#1}}}
\newcommand{\dmk}[1]{\textcolor{red!60!black}{#1}}
\newcommand{\papertitle}{Bottleneck Tokens for Unified Multimodal Retrieval}

\begin{document}

\hypersetup{pdftitle={\papertitle}}
\title{\papertitle}

\author{Siyu Sun\inst{1}, Jing Ren\inst{2}, Zhaohe Liao\inst{1}, Dongxiao Mao\inst{2}\\
Xiangyuan Ren\inst{2}, Yiyi Zhang\inst{1}, Haohua Zhao\inst{1}, Weixiong Lin\inst{2}\\
Jiang Shaohua\inst{2}, Liqing Zhang\inst{1}, Yuchao Zheng\inst{2}\thanks{Corresponding author.}}

\authorrunning{S.~Sun et al.}

\institute{Shanghai Jiao Tong University\\
\email{sunsiyu@sjtu.edu.cn}
\and
ByteDance\\
\email{zhengyuchao.yc@bytedance.com}}

\maketitle

\begin{abstract}
Adapting decoder-only multimodal large language models (MLLMs) for unified
multimodal retrieval faces two structural gaps.
First, existing methods rely on \emph{implicit pooling}, which overloads the
hidden state of a standard vocabulary token (\eg, \texttt{<EOS>}) as the
sequence-level representation---a mechanism never designed for information
aggregation.
Second, contrastive fine-tuning specifies \emph{what} the embedding should
match but provides no token-level guidance on \emph{how} information should be
compressed into it.
We address both gaps with two complementary components.
Architecturally, we introduce \textbf{Bottleneck Tokens (BToks)}, a small set
of learnable tokens that serve as a fixed-capacity explicit pooling mechanism.
For training, we propose \textbf{Generative Information Condensation}: a
next-token prediction objective coupled with a \textbf{Condensation Mask} that
severs the direct attention path from target tokens to query tokens.
All predictive signals are thereby forced through the BToks, converting the
generative loss into dense, token-level supervision for semantic compression.
At inference time, only the input and BToks are processed in a single forward
pass with negligible overhead over conventional last-token pooling.
On MMEB-V2 (78 datasets, 3 modalities, 9 meta-tasks), our approach achieves
state-of-the-art among 2B-scale methods under comparable data conditions, attaining an Overall score of
59.0 ({+3.6} over VLM2Vec-V2) with substantial gains on semantically
demanding tasks (\eg, {+12.6} on Video-QA).
\end{abstract}
\section{Introduction}

\begin{figure*}[t]
  \centering
  \includegraphics[width=\linewidth]{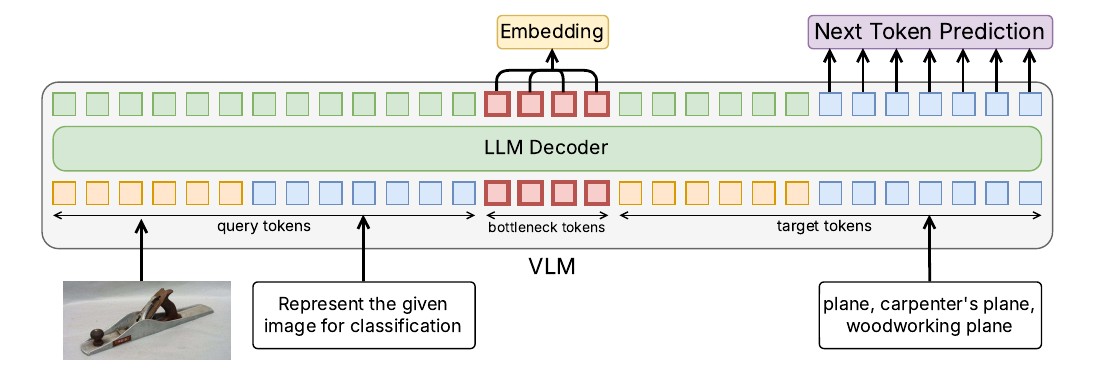}
  \caption{\textbf{Bottleneck tokens for unified multimodal retrieval.} The model uses a set of learnable bottleneck tokens (red) inserted between the query tokens and the target tokens in the LLM decoder. The query is embedded through these bottleneck tokens, while the model autoregressively predicts the target tokens. This structure enforces focused interaction between query and target, improving retrieval performance without altering the inference-time complexity.}
  \label{fig:sieve_arch}
\end{figure*}

Unified multimodal retrieval—encoding arbitrary sequences of interleaved
text and visual modalities (ranging from images to videos and visual
documents) into a shared embedding space—is a critical building block for
next-generation semantic search and RAG systems. Recent work has
demonstrated that decoder-only Multimodal Large Language Models (MLLMs)
can be effectively repurposed as universal embedding
models~\cite{jiang2024vlm2vec, jiang2024e5v}. These methods typically employ contrastive fine-tuning~\cite{oord2018cpc} and extract embeddings via
\textit{Implicit Pooling}, where the hidden state of the final token
(e.g., \texttt{<EOS>}) is treated as the global representation of the
input.

However, this reliance on implicit pooling introduces a fundamental
structural gap. In standard causal attention, the \texttt{<EOS>} token
is a vocabulary item repurposed for pooling—not a dedicated pooling
mechanism. While contrastive fine-tuning encourages the \texttt{<EOS>}
hidden state to capture matching-relevant features, it provides only a
pair-level training signal: the loss specifies \textit{what} the
embedding should match, but imposes no structural constraint on
\textit{how} information is aggregated across the input sequence. This
gap becomes particularly evident in semantically demanding tasks such as
QA-style retrieval, where the embedding must encode precise query intent
rather than surface-level visual similarity. Without token-level guidance
for the compression process, the model lacks a principled mechanism for
semantic condensation.

To address this gap, we propose a shift from implicit pooling to
\textit{Explicit Pooling}. We introduce \textbf{Bottleneck Tokens
(BToks)}, a set of learnable tokens designed solely as a dedicated
pooling mechanism for unified retrieval. Unlike the repurposed
\texttt{<EOS>} token, BToks are inserted into the
sequence to actively aggregate and condense visual-textual information.
By decoupling the pooling mechanism from the variable length of the
input, BToks provide a fixed-capacity bottleneck that standardizes
representation across diverse modalities.

However, introducing learnable tokens alone does not guarantee effective
condensation. Within the standard contrastive fine-tuning framework, we
further propose \textbf{Generative Information Condensation} as a
complementary training mechanism that supplies the token-level
supervision absent from pair-level contrastive loss. This mechanism
combines the \textbf{generative objective} (next-token prediction) with
a \textbf{Structural Attention Constraint} (or \textit{Condensation
Mask}). As illustrated in Fig.~\ref{fig:sieve_arch}, this mask severs the
direct attention path from the target text to the multimodal input—a
shortcut that would otherwise allow the model to minimize the prediction
loss by directly copying input features without condensing them into
BToks. Consequently, the target tokens are compelled to route their
attention solely through the BToks. This constraint converts the
generative objective into an embedding-aware condensation signal,
structurally forcing the model to compress all necessary semantic details
into the BToks to satisfy the prediction task.

We evaluate our approach on the comprehensive \textbf{MMEB-V2} benchmark
(78 datasets)~\cite{meng2025vlm2vecv2}. Results demonstrate that our method achieves
state-of-the-art performance among 2B-scale methods under comparable data conditions, with substantial
gains on semantically demanding tasks (e.g., +12.6 points on Video-QA
retrieval, +3.6 points on Image-QA) while maintaining full parity on
standard image-text retrieval tasks. This indicates that BToks
successfully capture fine-grained details without compromising general
semantic alignment, offering a robust solution for unified multimodal
retrieval.

Our contributions are summarized as follows:
\begin{itemize}
    \item \textbf{Architecture:} We introduce Bottleneck Tokens (BToks)
    as a dedicated explicit pooling mechanism that replaces the
    repurposed \texttt{<EOS>} token, addressing the structural
    limitations of implicit pooling in decoder-only MLLMs.
    \item \textbf{Mechanism:} We propose Generative Information
    Condensation, a training mechanism complementary to contrastive
    learning that combines next-token prediction with a structural
    attention constraint, providing dense, token-level supervision that
    compels the model to condense input semantics into BToks.
    \item \textbf{Results:} Our approach achieves state-of-the-art
    performance among 2B-scale models on MMEB-V2 under comparable
    conditions (public data, no external augmentation), delivering
    notable gains on semantically demanding tasks while preserving
    cross-modal consistency.
\end{itemize}

\section{Related Work}

\subsection{Multimodal Embedding Models}
\label{sec:rw-embedding}

\paragraph{Dual-encoder contrastive pretraining.}
CLIP~\cite{radford2021clip} and SigLIP~\cite{tschannen2025siglip2} established the paradigm of aligning independent image and text encoders via contrastive loss, optimizing both alignment and uniformity~\cite{wang2020alignment}. While effective for image-text matching, independent encoding precludes token-level cross-modal interaction, limiting performance on tasks requiring fine-grained understanding.

\paragraph{MLLM-based unified embedders.}
Decoder-only MLLMs~\cite{llavaov} have enabled unified
processing of interleaved text and visuals.
E5-V~\cite{jiang2024e5v} and VLM2Vec~\cite{jiang2024vlm2vec}
pioneered the adaptation of MLLMs for embedding via contrastive
fine-tuning.
Subsequent works have refined this paradigm through hard-negative
mining~\cite{lin2024mmembed, jian2025rzenembed}, multi-resolution
encoding~\cite{zhang2024gme}, and progressive
training~\cite{gu2025unime}.
Despite these improvements, they predominantly rely on
\textit{implicit pooling}—using the \texttt{<EOS>} hidden state as
the global representation—and contrastive fine-tuning as the sole
training signal.
Beyond single-vector embedding, alternative paradigms such as
multi-vector late interaction~\cite{faysse2024colpali} and
instruction-augmented retrieval~\cite{liu2024lamra} have also been
explored, though they target different efficiency-expressiveness
tradeoffs.

\paragraph{Reasoning-enhanced embeddings.}
Recent methods like TTE~\cite{cui2025ifmtte}, UME-R1~\cite{lan2025umer1}, and Embed-RL~\cite{jiang2026embed} leverage chain-of-thought reasoning to boost embedding quality.
While effective for complex queries, they rely on test-time reasoning overhead or external reinforcement learning stages.
In contrast, our work targets the core architectural bottleneck within the standard single-pass embedding framework.

\subsection{Training Objectives Beyond Contrastive Learning}
\label{sec:rw-objectives}

\paragraph{Contrastive-generative hybrids.}
Hybrid objectives have been explored to improve representation quality.
SLIP~\cite{mu2022slip}, CoCa~\cite{yu2022coca}, and CAFe~\cite{yu2025cafe} demonstrate that combining contrastive loss with captioning or generation objectives can enhance visual semantic alignment and reduce hallucination.
However, these methods use generation as an auxiliary signal for
feature enrichment or post-hoc embedding production, rather than
as a structural constraint on how information is compressed.
The generative loss operates alongside or after embedding
extraction, without being architecturally coupled with a dedicated
compression mechanism.

\subsection{Learnable Tokens for Representation Compression}
\label{sec:rw-compression}

\paragraph{Input-side compression.}
Learnable tokens are widely used as \textit{input-side} compressors
(e.g., Perceiver~\cite{jaegle2021perceiver},
Q-Former~\cite{li2023blip2},
Flamingo~\cite{alayrac2022flamingo}) to adapt visual features for
LLMs. These modules facilitate modality alignment but do not
produce retrieval embeddings.

\paragraph{Learnable tokens for embedding extraction.}
Conversely, \textit{output-side} learnable token approaches are
rarer in unified retrieval.
NV-Embed~\cite{lee2025nvembed} introduces a Latent Attention Layer
as an explicit replacement for \texttt{<EOS>} pooling.
While architecturally pioneering explicit pooling alternatives,
these methods rely exclusively on contrastive fine-tuning.
They lack a mechanism to structurally guide \textit{how}
information is compressed into the learnable tokens.
To our knowledge, our work is the first to integrate explicit
learnable pooling tokens with structurally constrained generative
supervision for unified multimodal retrieval. We term these tokens
\textbf{Bottleneck Tokens (BToks)} to reflect their role as a fixed-capacity information bottleneck for embedding extraction.
\section{Method}
\label{sec:method}

We address two complementary gaps in current MLLM-based embedding models.
At the \emph{structural} level, conventional designs pool a global representation from the hidden state of a borrowed vocabulary token (\eg, EOS)---a practice we term \textbf{implicit pooling}---rather than from a mechanism purpose-built for aggregation.
At the \emph{training-signal} level, the standard contrastive objective specifies \emph{which} query--candidate pairs should be close but provides no guidance on \emph{how} information should be compressed into the resulting embedding.
Our method resolves these two gaps with
(i)~\textbf{Bottleneck Tokens}, a small set of learnable tokens that \emph{explicitly} pool multimodal information into a fixed-width embedding (\cref{sec:btoks}), and
(ii)~\textbf{Generative Information Condensation}, a training mechanism that provides token-level supervision to guide \emph{how} information is condensed into those BToks (\cref{sec:gic}).
\Cref{sec:training} and \cref{sec:inference} describe the efficient training and inference procedures, respectively.

\paragraph{Notation.}
Let $\mathcal{Q}$ denote a set of queries and $\mathcal{C}$ a candidate pool.
For a query $q\in\mathcal{Q}$, the goal of retrieval is to rank candidates by relevance.
We measure relevance with cosine similarity $s(q,c)=\cos\!\bigl(e(q),\,e(c)\bigr)$ between embeddings $e(\cdot)\in\mathbb{R}^d$.
Throughout, we use a pre-trained multimodal large language model (MLLM) with $L$ transformer layers as the decoder-only backbone.

\paragraph{Baseline objective.}
Following standard practice, the backbone is fine-tuned with InfoNCE over in-batch negatives:
\begin{equation}
\label{eq:infonce}
  \mathcal{L}_{\mathrm{ctr}}
  \;=\;
  -\frac{1}{|\mathcal{B}|}\sum_{i\in\mathcal{B}}
    \log \frac{\exp\!\bigl(s(q_i,c_i^+)/\tau\bigr)}
              {\displaystyle\sum_{j\in\mathcal{B}}\exp\!\bigl(s(q_i,c_j)/\tau\bigr)},
\end{equation}
where $\tau$ is a fixed temperature and $c_i^+$ is the positive candidate for query $q_i$.
This loss tells the model \emph{what} to match but not \emph{how} to compress.

\subsection{Bottleneck Tokens}
\label{sec:btoks}

We introduce $K$ learnable \textbf{Bottleneck Tokens} (BToks) $B\in\mathbb{R}^{K\times d}$, appended after the input tokens to create an augmented sequence.
Given an input token sequence $x = [x_1,\dots,x_N]$, we construct:
\begin{equation}
\label{eq:btoks}
  B \;=\; [b_1,\dots,b_K], \quad b_k\in\mathbb{R}^d,
\end{equation}
\begin{equation}
\label{eq:augmented}
  \tilde{x} \;=\; [x_1,\dots,x_N,\;b_1,\dots,b_K].
\end{equation}
The augmented sequence $\tilde{x}$ is forwarded through all $L$ layers.
Let $h^{(L)}\!\bigl(\tilde{x}\bigr)$ denote the final-layer hidden states; we extract the $K$ positions corresponding to BToks:
\begin{equation}
\label{eq:hidden}
  h_b^{(L)} \;=\; \bigl[h_{N+1}^{(L)},\,\dots,\,h_{N+K}^{(L)}\bigr].
\end{equation}
The embedding is obtained by mean-pooling over these $K$ hidden states:
\begin{equation}
\label{eq:embed}
  e(x) \;=\; \mathrm{MeanPool}\!\bigl(h_b^{(L)}\bigr) \;=\; \frac{1}{K}\sum_{k=1}^{K} h_{N+k}^{(L)}.
\end{equation}
Mean pooling treats every BTok symmetrically, discouraging the model from collapsing information into a single token and encouraging each BTok to capture a complementary aspect of the input.
This replaces the implicit-pooling paradigm: instead of borrowing a vocabulary token (\eg, EOS) that was never designed for aggregation, BToks are purpose-built to aggregate information.

\subsection{Generative Information Condensation}
\label{sec:gic}

The contrastive loss $\mathcal{L}_{\mathrm{ctr}}$ operates on the final embedding $e(\cdot)$ and specifies \emph{which} pairs should be close, but provides no token-level guidance on \emph{how} information should be compressed into the BToks.
Generative Information Condensation fills this gap by combining a generative objective with a structural attention constraint that together provide direct, token-level supervision for the compression process.
The two components are coupled: without the structural constraint the generative loss is undermined by a shortcut (\cf below), while without the generative loss the constraint alone provides no learning signal.

\paragraph{Training sequence.}
During training, each sample is organized as a concatenation of three segments:
\begin{equation}
\label{eq:training_seq}
  \hat{x} \;=\; [\underbrace{x_1^{(q)},\dots,x_{N_q}^{(q)}}_{\text{query tokens}},\;
                  \underbrace{b_1,\dots,b_K}_{\text{BToks}},\;
                  \underbrace{x_1^{(t)},\dots,x_{N_t}^{(t)}}_{\text{target tokens}}].
\end{equation}
Note that $\hat{x}$ extends $\tilde{x}$ (Eq.~\ref{eq:augmented}) by appending the target tokens after the BToks.

\paragraph{Shortcut problem.}
A na\"{\i}ve approach would allow target tokens to attend to query tokens directly.
However, this creates a \emph{shortcut}: target tokens can simply copy information from the query rather than relying on the BToks, rendering the bottleneck ineffective and degrading embedding quality.

\paragraph{Condensation Mask.}
To eliminate this shortcut, we impose a \textbf{Structural Attention Constraint}, realized as a block-structured attention mask we call the \textbf{Condensation Mask}, over the three segments of $\hat{x}$:
\begin{equation}
\label{eq:mask}
  M \;=\;
  \begin{pmatrix}
    M_{\mathrm{causal}}^{(\mathrm{Q,Q})} & 0 & 0 \\[3pt]
    1 & M_{\mathrm{causal}}^{(\mathrm{B,B})} & 0 \\[3pt]
    0 & 1 & M_{\mathrm{causal}}^{(\mathrm{T,T})}
  \end{pmatrix},
\end{equation}
where rows correspond to the attending segment and columns to the attended segment.
Each entry indicates whether attention is permitted ($1$) or blocked ($0$);
$M_{\mathrm{causal}}^{(\cdot,\cdot)}$ denotes the standard causal (lower-triangular) mask within each segment.
The key properties are:
(a)~Query tokens attend only among themselves (causal);
(b)~BToks attend to all query tokens and causally among themselves, encouraging them to absorb query information;
(c)~Target tokens attend to all BToks and causally among themselves, but \emph{cannot} see any query token---eliminating the shortcut.
BToks attend causally among themselves---rather than bidirectionally---to remain compatible with standard causal-attention kernels (\eg, FlashAttention), which expect a single lower-triangular mask for the entire sequence.
Given the small number of BToks ($K\!\ll\!N$), we empirically find causal ordering among them sufficient.

\paragraph{Generative objective.}
Given the Condensation Mask, the target tokens can only access query information through the BToks.
We apply a standard next-token prediction (NTP) loss over the target segment:
\begin{equation}
\label{eq:ntp}
  \mathcal{L}_{\mathrm{ntp}}
  \;=\;
  -\frac{1}{N_t}\sum_{i=1}^{N_t}
    \log\,p_\theta\!\bigl(x_i^{(t)}\mid x_{<i}^{(t)},\,h_b^{(L)}\bigr),
\end{equation}
where $h_b^{(L)}$ denotes the BTok hidden states (Eq.~\ref{eq:hidden}).
Because the Condensation Mask blocks direct target-to-query attention, all query information available to the target segment is mediated exclusively through $h_b^{(L)}$.
This forces the model to compress all query information needed for generation into the BToks, providing direct, token-level supervision for the embedding.

\paragraph{Joint objective.}
The final training objective combines the inherited contrastive loss (Eq.~\ref{eq:infonce}) with the generative condensation loss:
\begin{equation}
\label{eq:joint}
  \mathcal{L} \;=\; \mathcal{L}_{\mathrm{ctr}} \;+\; \lambda\,\mathcal{L}_{\mathrm{ntp}},
\end{equation}
where $\mathcal{L}_{\mathrm{ctr}}$ is computed over the BToks mean-pooled embeddings $e(\cdot)$ (Eq.~\ref{eq:embed}) and $\lambda=\lambda(t)$ controls the strength of generative supervision (we use a step-dependent schedule detailed in the supplementary material).
$\mathcal{L}_{\mathrm{ctr}}$ specifies \emph{what} the embedding should match; $\mathcal{L}_{\mathrm{ntp}}$, enabled by the Condensation Mask, guides \emph{how} information is compressed into the BToks---making the two losses complementary rather than redundant.

\subsection{Efficient Training}
\label{sec:training}

Contrastive learning benefits from large batch sizes, yet each sample in our framework carries both a contrastive and a generative loss.

\paragraph{Sub-batch gradient cache.}
A global batch of $|\mathcal{B}|$ pairs is split into $S$ sub-batches that individually fit in GPU memory.
In \emph{Phase~A}, each sub-batch is forwarded to obtain BToks-pooled embeddings, which are buffered with their computation graphs detached.
After all $S$ forwards, the full-batch $\mathcal{L}_{\mathrm{ctr}}$ is computed on these buffered embeddings to obtain per-example gradients with respect to the embedding vectors.
In \emph{Phase~B}, each sub-batch is forwarded again: the cached embedding gradients are injected at the BTok pooling output to backpropagate $\mathcal{L}_{\mathrm{ctr}}$ through the full computation graph, while $\mathcal{L}_{\mathrm{ntp}}$ is computed and backpropagated concurrently in the same pass.
This decoupling allows the contrastive loss to see all $|\mathcal{B}|\times|\mathcal{B}|$ pairs while the generative loss remains local to each sub-batch, without requiring $|\mathcal{B}|$-sample activations to reside in memory simultaneously.
The Condensation Mask is realized via two standard causal attention calls compatible with FlashAttention; details are provided in the supplementary material.

\subsection{Inference}
\label{sec:inference}

At inference time, neither the target segment nor the generative loss is needed.
Each query or candidate is processed as $\tilde{x}=[x_1,\dots,x_N,b_1,\dots,b_K]$ (Eq.~\ref{eq:augmented}) with standard causal attention in a single forward pass.
The embedding $e(x)$ is obtained by mean-pooling the $K$ BTok hidden states (Eq.~\ref{eq:embed}), yielding a fixed-dimensional vector regardless of input length.
Compared to an EOS-pooling baseline, the only additional cost is processing $K$ extra tokens; for typical $K\!\ll\!N$ this overhead is negligible.
The resulting embedding is used for retrieval via cosine similarity, identically to conventional dense retrieval models.

\section{Experiments}
\label{sec:experiments}

We conduct comprehensive experiments on MMEB-V2~\cite{meng2025vlm2vecv2},
a large-scale multimodal embedding benchmark covering 78 datasets,
three modalities (Image, Video, and Visual Document),
and nine meta-tasks.
All results are reported as macro-averaged scores following the
official evaluation protocol.

\subsection{Experimental Setup}
\label{sec:exp_setup}

\paragraph{Model and training.}
We initialize BToks using the pre-trained EOS token embedding from Qwen2VL-2B-Instruct~\cite{wang2024qwen2},
a 2.2\,B-parameter vision--language model (VLM).
The number of bottleneck tokens is set to $K{=}4$ throughout unless
otherwise stated (see \cref{tab:ablation_k} for a sensitivity study).
Training uses the same publicly available data as
VLM2Vec-V2~\cite{meng2025vlm2vecv2}, comprising paired
image--text, video--text, and document--text samples,
ensuring a fair comparison.
We train for one epoch with the AdamW
optimizer at a learning rate of $5\!\times\!10^{-5}$
and a cosine schedule with linear warm-up.
The contrastive loss uses in-batch negatives with cross-GPU
gathering; the generative next-token prediction (NTP) loss shares
the same forward pass.
All experiments are conducted on NVIDIA A100-80\,GB GPUs.

\paragraph{Evaluation.}
We evaluate on the full MMEB-V2 benchmark.
Scores are macro-averaged first within each of the nine meta-tasks
and then within each modality group to produce three modality-level
scores (Image, Video, VisDoc) and a single Overall score.
This protocol accounts for the imbalanced number of datasets across
tasks and is consistent with the official leaderboard.

\paragraph{Baselines.}
We compare BToks with two groups of methods:
(i) \emph{2\,B-scale models trained under comparable settings}—%
ColPali~\cite{faysse2024colpali},
VLM2Vec-V1~\cite{jiang2024vlm2vec},
Gme~2.2\,B~\cite{zhang2024gme},
and our direct baseline VLM2Vec-V2~\cite{meng2025vlm2vecv2},
which is reproduced under identical data and training conditions;
and
(ii) \emph{reference methods} that operate under different conditions—%
either using larger backbones (${\ge}$\,7\,B), proprietary training
data, or chain-of-thought traces from external models.
The reference group includes
LamRA~\cite{liu2024lamra},
Gme~8.3\,B~\cite{zhang2024gme},
UniME-V2~\cite{gu2025unime},
CAFe~\cite{yu2025cafe},
UME-R1~\cite{lan2025umer1},
RzenEmbed-V1 and -V2~\cite{jian2025rzenembed},
and IFM-TTE~\cite{cui2025ifmtte}.

\subsection{Main Results}
\label{sec:main_results}
\begin{table*}[t]
\caption{\textbf{Main results on MMEB-V2 (78 datasets, 3 modalities, 9 meta-tasks).}
The upper block lists reference methods that use larger backbones ($\ge$\,7\,B),
proprietary training data $^\dagger$, or chain-of-thought traces
from external models $^\ddagger$.
The lower block compares 2B-scale models trained on publicly available data
without external augmentation.
VLM2Vec-V2 is reproduced under identical training settings as our direct baseline.
\textbf{Bold} and \underline{underline} denote the best and second-best results
in the 2B-scale block, respectively.}
    \label{tab:main_mmeb_v2}
    \centering
    \fontsize{8}{11}\selectfont
    \resizebox{\textwidth}{!}{%
    \setlength{\tabcolsep}{1pt}
    \begin{tabular}{l c c ccccc ccccc ccccc}
        \toprule
        \multirow{3}{*}{\textbf{Model}} & \multirow{3}{*}{\textbf{Size}} & \multirow{3}{*}{\textbf{Overall}}
        & \multicolumn{5}{c}{\textbf{Image} (acc@1)} & \multicolumn{5}{c}{\textbf{Video} (acc@1)} & \multicolumn{5}{c}{\textbf{VisDoc} (ndcg@5)} \\
        \cmidrule(lr){4-8} \cmidrule(lr){9-13} \cmidrule(lr){14-18}
        & & & Avg & CLS & QA & RET & GD & Avg & CLS & QA & RET & MRET & Avg & VDRv1 & VDRv2 & VR & OOD \\
        & & & {\scriptsize\color{gray}36} & {\scriptsize\color{gray}10} & {\scriptsize\color{gray}10} & {\scriptsize\color{gray}12} & {\scriptsize\color{gray}4} & {\scriptsize\color{gray}18} & {\scriptsize\color{gray}5} & {\scriptsize\color{gray}5} & {\scriptsize\color{gray}5} & {\scriptsize\color{gray}3} & {\scriptsize\color{gray}24} & {\scriptsize\color{gray}10} & {\scriptsize\color{gray}4} & {\scriptsize\color{gray}6} & {\scriptsize\color{gray}4} \\
        \midrule
        \multicolumn{18}{l}{\textit{\textbf{Reference: Larger Scale / Different Data Conditions}}} \\
        \midrule
        LamRA~\cite{liu2024lamra}             & 8.3B & 47.4 & 52.4 & 51.7 & 34.1 & 66.9 & 56.7 & 33.6 & 32.9 & 42.6 & 23.2 & 37.2 & 50.2 & 56.3 & 33.3 & 58.2 & 40.1 \\
        Gme~\cite{zhang2024gme}                           & 8.3B & 57.8 & 56.0 & 57.7 & 34.7 & 71.2 & 59.3 & 38.4 & 37.4 & 50.4 & 28.4 & 37.0 & 75.2 & 89.4 & 55.6 & 85.0 & 44.4 \\
        UniME-V2~\cite{gu2025unime}          & 8.0B & 59.6 & 71.8 & 65.6 & 68.7 & 73.1 & 90.9 & 39.0 & 37.2 & 50.6 & 28.9 & 39.6 & 56.7 & 61.8 & 42.0 & 70.5 & 37.9 \\
        CAFe~\cite{yu2025cafe}                  & 8.0B & 60.6 & 67.6 & 65.2 & 65.6 & 70.0 & 91.2 & 42.4 & 35.8 & 58.7 & 34.4 & 39.5 & 63.9 & 70.7 & 49.6 & 79.5 & 38.1 \\
        UME-R1$^\ddagger$~\cite{lan2025umer1}           & 8.3B & 64.1 & 71.3 & 67.1 & 69.2 & 71.9 & 84.9 & 47.5 & 48.6 & 60.7 & 38.2 & 39.3 & 65.7 & 75.7 & 50.5 & 83.7 & 29.2 \\
        RzenEmbed-V1$^\dagger$~\cite{jian2025rzenembed}  & 2.2B & 64.4 & 68.5 & 65.3 & 61.7 & 73.8 & 77.9 & 42.6 & 45.6 & 47.5 & 38.3 & 36.7 & 74.4 & 87.0 & 57.6 & 85.4 & 43.3 \\
        RzenEmbed-V2$^\dagger$~\cite{jian2025rzenembed}  & 8.3B & 71.6 & 75.9 & 70.6 & 71.7 & 78.5 & 92.1 & 55.7 & 58.8 & 63.5 & 51.0 & 45.5 & 77.1 & 89.7 & 60.7 & 88.7 & 44.4 \\
        IFM-TTE$^\ddagger$~\cite{cui2025ifmtte}         & 8.3B & 73.1 & 77.9 & 76.7 & 78.5 & 74.6 & 89.3 & 59.2 & 60.5 & 67.9 & 51.7 & 54.9 & 76.2 & 85.2 & 71.5 & 92.8 & 33.7 \\
        \midrule
        \multicolumn{18}{l}{\textit{\textbf{2B Scale (Comparable Settings: Public Data, No External Augmentation)}}} \\
        \midrule
        ColPali~\cite{faysse2024colpali}                   & 2.9B & 44.4 & 34.9 & 40.3 & 11.5 & 48.1 & 40.3 & 28.2 & 26.7 & 37.8 & 21.6 & 25.5 & \underline{71.0} & \underline{83.6} & \underline{52.0} & 81.1 & \textbf{43.1} \\
        VLM2Vec-V1~\cite{jiang2024vlm2vec}               & 2.2B & 47.0 & 59.7 & 58.7 & 49.3 & 65.0 & 72.9 & 28.6 & 33.4 & 30.5 & 20.6 & 30.8 & 41.6 & 49.8 & 13.5 & 51.8 & 33.6 \\
        Gme~\cite{zhang2024gme}                           & 2.2B & 54.1 & 51.9 & 56.9 & 41.2 & \underline{67.8} & 53.4 & \underline{33.6} & 34.9 & \underline{42.0} & 25.6 & 31.1 & \textbf{72.7} & \textbf{86.2} & \textbf{54.0} & \textbf{82.5} & \textbf{43.1} \\
        VLM2Vec-V2~\cite{meng2025vlm2vecv2}             & 2.2B & \underline{55.4} & \underline{64.2} & \textbf{64.5} & \underline{56.2} & 67.2 & \underline{74.7} & \underline{33.6} & \underline{39.1} & 34.4 & \underline{28.2} & \underline{32.0} & 58.5 & 66.0 & 37.1 & 77.6 & 32.4 \\
        \rowcolor{gray!15}
        \textbf{BToks (Ours)}                    & \textbf{2.2B} & \textbf{59.0} & \textbf{66.0} & \underline{64.3} & \textbf{59.8} & \textbf{68.8} & \textbf{77.4} & \textbf{39.9} & \textbf{43.7} & \textbf{47.0} & \textbf{33.0} & \textbf{33.6} & 62.7 & 71.1 & 38.6 & \underline{81.3} & 38.1 \\
        \emph{vs.\ VLM2Vec-V2}                  & --   & \cmark{+3.6} & \cmark{+1.8} & \dmk{-0.2} & \cmark{+3.6} & \cmark{+1.6} & \cmark{+2.7} & \cmark{+6.3} & \cmark{+4.6} & \cmark{+12.6} & \cmark{+4.8} & \cmark{+1.6} & \cmark{+4.2} & \cmark{+5.1} & \cmark{+1.5} & \cmark{+3.7} & \cmark{+5.7} \\
        \bottomrule
    \end{tabular}
    }%
\end{table*}

\cref{tab:main_mmeb_v2} presents the main results.
Among 2\,B-scale models trained on publicly available data,
BToks achieves an Overall score of \textbf{59.0},
surpassing VLM2Vec-V2 by \textbf{+3.6} points.
The improvements are consistent across all three modality groups:
Image (+1.8 over VLM2Vec-V2),
Video (+6.3),
and VisDoc (+4.2).

\paragraph{Image tasks.}
BToks scores 66.0 on the Image group (+1.8 over VLM2Vec-V2),
which aggregates classification, QA, retrieval, and
grounding tasks.
The gain is concentrated in semantically demanding sub-tasks:
Image-QA improves by \textbf{+3.6} (59.8 vs.\ 56.2) and
Grounding by +2.7 (77.4 vs.\ 74.7),
while Classification remains on par ($-$0.2).
This pattern is consistent with our hypothesis that
generative information condensation encourages the
bottleneck tokens to preserve fine-grained query--target
correspondence rather than surface-level visual features.

\paragraph{Video tasks.}
Video retrieval has been a persistent bottleneck for
VLM-based embedding models due to the long token sequences
involved.
BToks achieves 39.9 on the Video group, outperforming all
2\,B-scale methods by a wide margin
(VLM2Vec-V2: 33.6, Gme: 33.6).
The most notable gain appears on Video-QA
(\textbf{+12.6}, 47.0 vs.\ 34.4),
where the model must compress lengthy temporal contexts
into an embedding that captures precise question intent.
We attribute this to the fixed-capacity bottleneck:
BToks forces the model to distill discriminative details
from variable-length frame sequences rather than
distributing information diffusely across the
\texttt{<EOS>} hidden state.

\paragraph{Visual document tasks.}
BToks achieves 62.7 on the VisDoc group,
a gain of \textbf{+4.2} over VLM2Vec-V2 (58.5).
Document understanding requires jointly processing layout,
textual content, and visual elements---information that
is spatially dispersed across the input sequence.
The condensation mask ensures that all such information
must flow through the bottleneck tokens,
preventing the model from relying on sparse positional
shortcuts and encouraging holistic document-level
compression.

\paragraph{Comparison with reference methods.}
Despite using a backbone four times smaller and only publicly available
training data, BToks (59.0) surpasses Gme~8.3\,B (57.8) and closely
approaches UniME-V2~8.0\,B (59.6) and CAFe~8.0\,B (60.6)---all of
which employ $\geq$7\,B backbones.
Methods that additionally leverage proprietary data
(RzenEmbed-V2~8.3\,B, 71.6) or external chain-of-thought supervision
(IFM-TTE~8.3\,B, 73.1) achieve substantially higher absolute scores;
however, these advantages are orthogonal to---and potentially
combinable with---our approach, as BToks modifies only the pooling and
training mechanism without constraining the data pipeline.

\subsection{Ablation Studies}
\label{sec:ablation}

\subsubsection{Component Ablation}
\label{sec:ablation_components}
\begin{table}[t]
\centering
\caption{\textbf{Ablation study on key components} evaluated on MMEB-V2~\cite{meng2025vlm2vecv2}.
  Starting from the full BToks framework (first row),
  we remove one component at a time:
  (i) replacing bottleneck tokens with last-token (EOS) pooling,
  (ii) removing the Condensation Mask while retaining bottleneck tokens and the generative objective,
  (iii) removing the generative objective (NTP loss) while keeping all other components.
  \textbf{Bold} and \underline{underline} denote the best and second-best, respectively.}
\label{tab:ablation_components}
\resizebox{0.9\columnwidth}{!}{%
\setlength{\tabcolsep}{10pt}
\renewcommand{\arraystretch}{1.05}
\begin{tabular}{lcccc}
\toprule
\textbf{Variant} & \textbf{Overall} & \textbf{Image} & \textbf{Video} & \textbf{VisDoc} \\
\midrule
\rowcolor{gray!10}
BToks (Full) & \textbf{59.0} & \textbf{66.0} & \textbf{39.9} & \textbf{62.7} \\
EOS pooling & 56.1 & 63.8 & 37.1 & 58.9 \\
w/o Condensation Mask & \underline{57.9} & 64.5 & \underline{39.8} & \underline{61.7} \\
w/o Generative Objective & 57.6 & \underline{65.6} & \underline{39.8} & 59.1 \\
\bottomrule
\end{tabular}
}
\end{table}

\cref{tab:ablation_components} isolates the contribution of
each component by removing one at a time from the full BToks
framework.

\paragraph{Bottleneck tokens.}
Replacing BToks with standard EOS pooling
(row~2 vs.\ row~1) reduces Overall from 59.0 to 56.1
($-$2.9).
The degradation is broadly distributed across modalities:
Image drops by 2.2, Video by 2.8, and VisDoc by 3.8.
The Video and VisDoc declines are particularly notable,
as both modalities involve long or layout-rich input
sequences where a single \texttt{<EOS>} state must
compress substantially more information than in
image-only tasks.
This confirms that explicit fixed-capacity pooling is
critical when the token-to-embedding compression ratio
is high.

\paragraph{Condensation mask.}
Removing the condensation mask while keeping BToks and
the generative objective (row~3) lowers Overall by 1.1
(57.9 vs.\ 59.0).
The loss is concentrated on Image ($-$1.5) and VisDoc
($-$1.0), whereas Video is virtually unchanged ($-$0.1).
Without the mask, target tokens can directly attend to
query tokens, allowing the generative loss to be
satisfied through shortcut attention paths that bypass
the bottleneck.

\paragraph{Generative objective.}
Removing the next-token prediction loss while retaining
BToks and the condensation mask (row~4) causes a larger
Overall drop of 1.4 (57.6 vs.\ 59.0).
The effect is strikingly modality-specific:
VisDoc suffers the steepest decline ($-$3.6),
Image sees only a marginal change ($-$0.4),
and Video is again nearly unaffected ($-$0.1).
We attribute the VisDoc sensitivity to the token-level
nature of the NTP objective: documents contain dense
textual content whose sequential structure is directly
amenable to generative supervision, so removing it
eliminates the strongest compression signal for this
modality.

\paragraph{Joint effect.}
The condensation mask and the generative objective
form a synergistic pair that we term
\emph{generative information condensation}.
Removing the mask alone costs 1.1 points
and removing the objective alone costs 1.4;
the combined loss when switching to EOS pooling (which removes
both, along with the bottleneck tokens themselves) is 2.9,
exceeding the sum of the individual losses (2.5).
This super-additive degradation confirms that the two components
are complementary: the mask forces information through the
bottleneck, and the generative objective ensures that
the resulting representation is sufficiently rich.

\subsubsection{Number of Bottleneck Tokens}
\label{sec:ablation_k}
\begin{table}[t]
\centering
\caption{Effect of the number of bottleneck tokens $K$ on MMEB-V2~\cite{meng2025vlm2vecv2}.
  We report macro-averaged scores across all 78 datasets (Overall) and
  per-modality groups (Image, Video, VisDoc).
  \textbf{Bold} and \underline{underline} denote the best and second-best, respectively.
  The highlighted column ($K{=}4$) is our default setting.}
\label{tab:ablation_k}
\resizebox{0.8\columnwidth}{!}{%
\setlength{\tabcolsep}{8pt}
\begin{tabular}{lcc>{\columncolor{gray!10}}cccc}
\toprule
$K$ & 1 & 2 & 4 & 8 & 16 & 32 \\
\midrule
Overall  & 57.15 & 57.22 & \textbf{58.96} & \underline{58.34} & 58.32 & 57.99 \\
\midrule
Image    & 65.40 & 65.62 & \underline{65.97} & 65.82 & \textbf{66.03} & 65.86 \\
Video    & 37.22 & 37.86 & \textbf{39.94} & \underline{39.31} & 38.77 & 38.00 \\
VisDoc   & 59.71 & 59.14 & \textbf{62.72} & 61.39 & \underline{61.41} & 61.19 \\
\bottomrule
\end{tabular}
}
\end{table}

\cref{tab:ablation_k} reports performance as the number of
bottleneck tokens $K$ varies from 1 to 32.
Overall performance peaks at $K{=}4$ (58.96) and
degrades on both sides: $K{=}1$ scores 57.15 ($-$1.81)
and $K{=}32$ scores 57.99 ($-$0.97).

The three modalities exhibit different sensitivities.
\emph{Image} is relatively stable, ranging from 65.40 ($K{=}1$)
to 66.03 ($K{=}16$), with the difference between $K{=}4$ (65.97)
and the best barely 0.06 points.
\emph{Video} is the most sensitive, peaking sharply at $K{=}4$
(39.94) and declining to 38.00 at $K{=}32$—a 1.94-point drop
that indicates over-parameterised bottlenecks fragment temporal
information across too many tokens.
\emph{VisDoc} follows a similar trend to Overall, with $K{=}4$
(62.72) clearly ahead of $K{=}8$ (61.39) and beyond.

The consistent optimality of $K{=}4$ across modalities supports
our default setting and highlights a favorable trade-off:
each additional token adds negligible inference overhead
(see \cref{tab:efficiency}), yet too many tokens undermine the
compression pressure that drives the bottleneck mechanism.

\subsection{Cross-Modal Generalization}
\label{sec:cross_modal}
\begin{table}[t]
\centering
\caption{Performance comparison between BToks and the w/o-BToks baseline when models are trained on single-modality data (Image / Video / VisDoc). Numbers in parentheses denote absolute improvements over the corresponding w/o-BToks baseline. The results show that BToks more effectively leverages the same training data and yields robustness and cross-modal generalization gains under data-scarce conditions.}
\label{tab:modality}
\resizebox{0.8\columnwidth}{!}{%
\setlength{\tabcolsep}{8pt}
\renewcommand{\arraystretch}{1.05}
\begin{tabular}{lcccc}
\toprule
\textbf{Modality} & \textbf{Image} & \textbf{Video} & \textbf{VisDoc} & \textbf{Overall} \\
\midrule
Datasets & 36 & 18 & 24 & 78 \\
\midrule
Image (w/o BToks) & 64.61 & 33.18 & 31.15 & 47.06 \\
\rowcolor{gray!10}
Image (BToks) &
  \makecell[c]{\textbf{65.39} \\ \scriptsize(+0.78)} &
  \makecell[c]{\textbf{39.42} \\ \scriptsize(+6.24)} &
  \makecell[c]{\textbf{47.03} \\ \scriptsize(+15.88)} &
  \makecell[c]{\textbf{53.75} \\ \scriptsize(+6.69)} \\
\midrule
Video (w/o BToks) & 27.00 & 28.00 & 47.43 & 33.79 \\
\rowcolor{gray!10}
Video (BToks) &
  \makecell[c]{\textbf{30.92} \\ \scriptsize(+3.92)} &
  \makecell[c]{\textbf{31.45} \\ \scriptsize(+3.45)} &
  \makecell[c]{\textbf{48.19} \\ \scriptsize(+0.76)} &
  \makecell[c]{\textbf{36.35} \\ \scriptsize(+2.56)} \\
\midrule
VisDoc (w/o BToks) & 44.96 & 26.61 & 53.54 & 43.37 \\
\rowcolor{gray!10}
VisDoc (BToks) &
  \makecell[c]{\textbf{50.89} \\ \scriptsize(+5.93)} &
  \makecell[c]{\textbf{32.47} \\ \scriptsize(+5.86)} &
  \makecell[c]{\textbf{59.60} \\ \scriptsize(+6.06)} &
  \makecell[c]{\textbf{49.32} \\ \scriptsize(+5.95)} \\
\bottomrule
\end{tabular}
}
\end{table}

To understand whether the bottleneck tokens capture
modality-agnostic semantic structure,
\cref{tab:modality} evaluates models trained on a
\emph{single} modality and tested across all three.

\paragraph{Image-only training.}
When trained only on Image data, BToks achieves an
Overall score 6.69 points higher than the baseline
without bottleneck tokens.
The most striking result is the transfer to VisDoc:
BToks gains +15.88 over the baseline,
despite never having seen any document data during training.
This indicates that the bottleneck representations learned
from images generalize to the layout and textual cues present
in visual documents.

\paragraph{Video-only training.}
Video-only BToks improves Overall by +2.56.
The smaller gain relative to Image-only training reflects the
inherent domain gap between video and the two other modalities,
yet the improvement is nonetheless consistent across all
evaluation groups.

\paragraph{VisDoc-only training.}
Training exclusively on VisDoc data yields
an Overall gain of +5.95 for BToks.
Notably, the VisDoc-trained bottleneck tokens transfer well
to Image tasks, confirming bidirectional
cross-modal transferability.

\subsection{Efficiency Analysis}
\label{sec:efficiency}
\begin{table}[t]
\centering
\caption{\textbf{Training and inference efficiency.}
  BToks achieves superior performance under comparable efficiency.
  We report training cost (GPU$\cdot$hours on A100-80G),
  inference latency (p50/p90/mean in ms), and throughput (queries/s)
  on MMEB-V2.}
\label{tab:efficiency}
\resizebox{\columnwidth}{!}{%
\setlength{\tabcolsep}{4pt}
\begin{tabular}{lccccc}
\toprule
\textbf{Method} & \textbf{Train (GPU$\cdot$h)} & \textbf{p50 (ms)} & \textbf{p90 (ms)} & \textbf{Mean (ms)} & \textbf{Thrpt (/s)} \\
\midrule
VLM2Vec-V2~\cite{meng2025vlm2vecv2} & 504 & 74.90 & 87.10 & 79.09 & 50.57 \\
\rowcolor{gray!10}
\textbf{BToks (Ours)} & 695 & 76.48 & 84.73 & 80.06 & 49.96 \\
\midrule
\multicolumn{6}{@{}l}{\footnotesize
  \textit{Overhead of bottleneck tokens:}
  \quad
  mean $+$0.97\,ms ($+$1.2\%),\quad
  throughput $-$0.61/s ($-$1.2\%).} \\
\bottomrule
\end{tabular}
}
\end{table}

\cref{tab:efficiency} compares the training and inference
efficiency of BToks against its direct baseline VLM2Vec-V2,
measured under identical hardware and input conditions
(batch size 1, image resolution $384{\times}384$, sequence
length 1024).

\paragraph{Training cost.}
BToks requires 695 GPU$\cdot$hours on A100-80\,GB,
compared to 504 for VLM2Vec-V2—an increase of 37.9\%.
This additional cost arises from the generative NTP loss,
which shares the forward pass but adds a decoding head
and gradient computation for the next-token prediction branch.
We note that training is a one-time cost; at deployment,
the generative head is discarded.

\paragraph{Inference latency.}
At inference time, BToks introduces minimal overhead.
The p50 latency increases by only 1.58\,ms (+2.1\%),
the mean latency by 0.97\,ms (+1.2\%),
and throughput decreases by 0.61 queries/s ($-$1.2\%).
The p90 latency actually \emph{decreases} from 87.10\,ms
to 84.73\,ms, suggesting that the fixed-size bottleneck
output reduces variance in the final pooling and projection
steps.

\paragraph{Practical implications.}
The near-parity inference profile is a direct consequence of
our design: the $K{=}4$ bottleneck tokens are simply mean-pooled
into a single embedding vector of the same dimensionality as
the baseline, requiring no additional modules at test time.
The training-time overhead is a one-off investment that yields
a +3.6-point improvement in Overall, a favorable cost--performance
trade-off for production retrieval systems.

\section{Conclusion}
\label{sec:conclusion}

We have presented BToks, a method that introduces a small set
of learnable bottleneck tokens into a vision--language model
to serve as the explicit pooling mechanism for multimodal
embedding extraction.
By appending $K{=}4$ bottleneck tokens to every input sequence
and mean-pooling their final-layer representations,
BToks replaces the implicit last-token (EOS) pooling convention
with a dedicated, structurally constrained aggregation pathway.

Two complementary training-time mechanisms—collectively termed
\emph{generative information condensation}—ensure that the
bottleneck tokens capture rich, well-compressed representations:
(i)~a \emph{condensation mask} that restricts attention so that
only the bottleneck tokens may attend to all input positions,
forcing information to flow through them;
and (ii)~a \emph{generative objective} (next-token prediction)
that provides dense, token-level supervision beyond the
sparse pair-level signal of contrastive learning.
Both mechanisms operate exclusively during training and add
no overhead at inference time, where only the bottleneck token
embeddings are retained.

Experiments on MMEB-V2 (78 datasets, 3 modalities, 9 meta-tasks)
demonstrate that BToks achieves an Overall score of 59.0 with
a 2.2\,B-parameter backbone, outperforming VLM2Vec-V2 by +3.6
points under identical training data and settings.
The gains are consistent across Image (+1.8), Video (+6.3),
and Visual Document (+4.2) modalities.
Ablation studies confirm that each component contributes
meaningfully, with the bottleneck tokens themselves providing
the largest individual gain ($-$2.9 when removed) and the
condensation mask and generative objective exhibiting a
super-additive interaction.
Single-modality training experiments further reveal strong
cross-modal transfer, particularly a +15.88-point improvement
on Visual Document tasks when trained only on Image data,
indicating that the bottleneck representations are
modality-agnostic.
These improvements come at negligible inference cost:
p50 latency increases by only 1.58\,ms (+2.1\%) and throughput
decreases by merely 1.2\%.

\paragraph{Limitations and future work.}
Our study has several limitations that suggest directions for
future research.
First, all experiments use a single 2.2\,B backbone
(Qwen2VL-2B); scaling BToks to larger VLMs
(e.g., 7--8\,B) would clarify whether the bottleneck mechanism
remains effective when the base model already has greater
representational capacity, and could close the gap with
reference methods that rely on larger backbones or proprietary
data.
Second, the optimal number of bottleneck tokens $K{=}4$
is determined empirically; a learnable or input-adaptive $K$
could improve flexibility for inputs of varying complexity.
Third, the current generative objective uses standard
next-token prediction; exploring alternative dense supervision
signals—such as masked reconstruction or retrieval-augmented
generation—may further strengthen the condensation process.
Finally, while we demonstrate cross-modal transfer in
single-modality settings, extending BToks to modalities beyond
vision and text (e.g., audio, 3-D point clouds) remains an
open and promising direction.

\bibliographystyle{splncs04}
\bibliography{main}

\clearpage
\appendix
\renewcommand{\thesection}{\Alph{section}}
\renewcommand{\theHsection}{appendix.\Alph{section}}
\renewcommand{\theHsubsection}{appendix.\Alph{section}.\arabic{subsection}}

\section{Experimental Details}
\label{sec:exp_details}

This section provides implementation details, training data composition, and hyper-parameter settings that complement the experimental setup in \cref{sec:exp_setup}.

\subsection{Backbone and Implementation}
We instantiate BToks on top of the Qwen2VL-2B-Instruct backbone used by VLM2Vec-V2~\cite{meng2025vlm2vecv2}, treating all text and visual tokens as a single causal sequence. Images and video frames are encoded into patch tokens by the frozen vision tower, while text is tokenized with the Qwen2VL tokenizer. Following the formulation in \cref{sec:btoks}, we append $K$ trainable BToks after the input tokens and forward the augmented sequence through the Transformer.

BToks are global learnable embeddings shared across all examples, initialized from the
end-of-sequence (\texttt{<EOS>}) embedding and optimized jointly with the trainable
adapters in the backbone. At the final layer, we mean-pool the hidden
states corresponding to BToks to obtain a fixed-dimensional embedding
$e(x)$, which is used both for the contrastive loss and for downstream
retrieval evaluation. Unless otherwise stated, we use $K{=}4$, which we
empirically find to provide the best trade-off between capacity and
regularization.

For parameter-efficient training, we only update low-rank adapter
(LoRA) weights and BToks while keeping the remaining backbone
parameters frozen (details in \cref{sec:hyperparams}). The same
architecture and retrieval interface are used across all modalities
(Image, Video, VisDoc), so that BToks serve as a unified drop-in
pooling mechanism for any single-encoder multimodal LLM.

\subsection{Training Data and Sampling}
\label{sec:training_data}

All experiments use MMEB-V2, a large-scale benchmark for general-purpose
multimodal embeddings. It contains 78 datasets across nine meta-tasks
covering image, video, and visual-document retrieval. We follow the
official configuration and splits released by the benchmark authors,
including the canonical metric for each dataset (e.g., accuracy, Hit@1,
or NDCG@5).

For training, we reuse the MMEB-V2-style data mixture introduced by
VLM2Vec\allowbreak-V2. The mixture combines:
\begin{itemize}
  \item image-text datasets from MMEB train covering classification,
        visual question answering, retrieval, and grounding;
  \item video datasets such as LLaVA-Hound style video captioning and
        video QA, which we convert into retrieval-style pairs by
        treating captions and videos as queries/targets in both
        directions;
  \item visual-document datasets from ViDoRe and VisRAG that support
        document-level retrieval and question answering over structured
        pages.
\end{itemize}
We adopt the same prompt templates, instruction-style formatting, and
per-dataset sampling weights as in VLM2Vec-V2 so that the only
differences between BToks and the reproduced baseline are the
bottlenecked retrieval interface and the training objectives.

\Cref{tab:training_data} summarizes the training data composition by
modality and source.

\begin{table}[t]
  \centering
  \footnotesize
  \caption{Summary of the training data mixture. All data sources follow
  the VLM2Vec-V2 configuration. Weights are relative sampling weights
  within each training batch.}
  \label{tab:training_data}
\setlength{\tabcolsep}{4pt}
  \begin{tabular}{llcrr}
    \toprule
    Modality & Dataset & Source & Weight & \# Samples \\
    \midrule
    \multirow{20}{*}{Image}
    & ImageNet-1K        & \multirow{20}{*}{MMEB~\cite{jiang2024vlm2vec}}  & 1.0 & 100,000 \\
    & N24News            &   & 1.0 &  48,988 \\
    & HatefulMemes       &   & 0.5 &   8,500 \\
    & VOC2007            &   & 0.5 &   7,844 \\
    & SUN397             &   & 0.5 &  19,850 \\
    & OK-VQA             &   & 0.5 &   9,009 \\
    & A-OKVQA            &   & 0.5 &  17,056 \\
    & DocVQA             &   & 1.0 &  39,463 \\
    & InfographicsVQA    &   & 0.5 &  23,946 \\
    & ChartQA            &   & 0.5 &  28,299 \\
    & Visual7W           &   & 1.0 &  69,817 \\
    & VisDial            &   & 1.0 & 123,287 \\
    & VisualNews (t2i)   &   & 1.0 &  99,903 \\
    & VisualNews (i2t)   &   & 1.0 & 100,000 \\
    & MSCOCO (t2i)       &   & 1.0 & 100,000 \\
    & MSCOCO (i2t)       &   & 1.0 & 113,287 \\
    & MSCOCO (grounding) &   & 1.0 & 100,000 \\
    & CIRR               &   & 0.5 &  26,116 \\
    & NIGHTS             &   & 0.5 &  15,941 \\
    & WebQA              &   & 0.5 &  17,166 \\
    \midrule
    \multirow{3}{*}{Video}
    & Video Caption (t2v)  & \multirow{3}{*}{LLaVA-Hound~\cite{zhang2024llavahound}} & 5.0 & 301,751 \\
    & Video Caption (v2t)  & & 5.0 & 301,751 \\
    & Video QA              & & 5.0 & 255,000 \\
    \midrule
    \multirow{2}{*}{VisDoc}
    & VisRAG (in-domain)   & VisRAG~\cite{yu2024visrag}       & 12.0 & 122,752 \\
    & ColPali Train        & ColPali~\cite{faysse2024colpali} & 10.0 & 118,195 \\
    \midrule
    \multicolumn{3}{l}{\textbf{Total (25 datasets)}} & \textbf{52.0} & \textbf{2,167,921} \\
    \bottomrule
  \end{tabular}
\end{table}

For the modality-specific variants (Image-only, Video-only,
VisDoc-only) discussed in \cref{sec:cross_modal}, we restrict the training
data to the corresponding subset of the above mixture while keeping
the backbone, BToks configuration, and training budget identical to the
unified model. This allows us to isolate the effect of BToks under
specialized training and to study cross-modal generalization.

\subsection{Optimization and Hyper-parameters}
\label{sec:hyperparams}

We follow the MMEB-V2 training recipe used by VLM2Vec\allowbreak-V2 and
IFM\allowbreak-TTE~\cite{cui2025ifmtte}. We use
Qwen2VL\allowbreak-2B-Instruct as the
backbone, with a global batch size of $1{,}024$ implemented via
gradient caching over sub-batches of size $64$. The main models are
trained for $5{,}000$ optimization steps, which corresponds to roughly
one epoch over the training mixture. We adopt AdamW with a peak
learning rate of $5\!\times\!10^{-5}$ and cosine decay with linear
warm-up; weight decay matches the reproduced VLM2Vec\allowbreak-V2
baseline. The
temperature in the InfoNCE loss is fixed to $0.02$.

\paragraph{Parameter-efficient fine-tuning.}
We use LoRA with rank $r{=}16$ (see \cref{tab:lora_rank_sensitivity} for a sensitivity study). Adapters are inserted into all self-attention and MLP layers; only LoRA parameters and BToks are updated while the remaining backbone weights stay frozen.

\paragraph{Joint objective schedule.}
BToks introduces a joint training objective—realizing the Generative Information Condensation mechanism described in \cref{sec:gic}—that combines contrastive retrieval and next-token prediction (NTP) under the Condensation Mask:
\begin{equation}
  \mathcal{L}(t)
  = \mathcal{L}_{\mathrm{ctr}} +
    \lambda(t)\,\mathcal{L}_{\mathrm{ntp}},
\end{equation}
where $t$ denotes the optimization step. In all main experiments, we use a simple two-stage schedule for $\lambda(t)$: for the first $2{,}000$ steps (${\approx}40\%$ of training) we set $\lambda(t)=0.1$ to inject generative supervision into the bottleneck interface, and for the remaining $3{,}000$ steps we set $\lambda(t)=0$, effectively continuing training with a pure contrastive objective over BToks. This schedule lets the model first learn to route predictive signals through BToks and then focus on refining the embedding space without further NTP updates.

Unless otherwise noted, all headline results in
\cref{sec:main_results} use this default configuration: a
Qwen2VL\allowbreak-2B-Instruct backbone, LoRA adapters, a global batch size of
$1{,}024$, and the two-stage $\lambda$ schedule above.
\Cref{sec:sensitivity} further reports sensitivity studies on the LoRA
rank and the $\lambda(t)$ schedule.

\section{Two-pass KV Cache Implementation}
\label{sec:two_pass_kv}

\Cref{sec:gic} defines the Condensation Mask (\cref{eq:mask}), a
block-structured attention constraint that routes all
query-to-target information through BToks. A na\"{\i}ve implementation
would require a dense mask that cannot leverage optimized causal
kernels such as FlashAttention. We therefore use a \emph{two-pass}
implementation that realizes the same constraint with standard causal
attention while remaining fully differentiable.

\paragraph{Two-pass formulation.}
We decompose the masked forward into:
(i) a pass over query tokens and BToks to compute BTok representations and
their KV cache; and (ii) a pass over target tokens that treats the BTok KV
cache as a fixed prefix, so that target tokens can attend to BToks but never
to query tokens.
Because the cached KV remains in the autograd graph, gradients from the
next-token prediction (NTP) loss still backpropagate through BToks into the
query side.

\begin{algorithm}[t]
    \caption{Condensation Mask via two-pass KV cache (training only)}
    \label{alg:two_pass_kv}
    \begin{algorithmic}[1]
        \Require
        Query tokens \(x^{(q)}\), Bottleneck Tokens \(B = [b_1,\dots,b_K]\),
        target tokens \(x^{(t)}\), decoder depth \(L\)
        \Ensure
        Retrieval embedding \(e(x)\) from BToks and NTP loss \(\mathcal{L}_{\mathrm{ntp}}\)

        \State \textbf{// Pass 1: query + BToks}
        \State \(s^{(1)} \gets \mathrm{concat}(x^{(q)}, B)\)
        \State \(h^{(1)}_0 \gets \mathrm{Embed}(s^{(1)})\)
        \For{\( \ell = 1 \) \textbf{to} \( L \)}
            \State
            \(\big(h^{(1)}_\ell, \text{KV}^{(1)}_\ell\big)
                \gets \mathrm{DecoderLayer}_\ell\big(
                    h^{(1)}_{\ell-1};
                    \text{mask} = \text{causal},
                    \text{use\_cache} = \texttt{True}
                \big)\)
            \State Extract KV at BTok positions:
            \(\text{KV}^{B}_\ell \gets \mathrm{SelectBTokKV}\big(\text{KV}^{(1)}_\ell\big)\)
        \EndFor
        \State BTok hidden states at the last layer:
        \(h_b^{(L)} \gets \mathrm{SelectBTokHidden}\big(h^{(1)}_L\big)\)
        \State Retrieval embedding:
        \(e(x) \gets \mathrm{MeanPool}\big(h_b^{(L)}\big)\)
        \State \textbf{// Note: } \(\text{KV}^{B}_\ell\) keeps full autograd connectivity

        \State \textbf{// Pass 2 (training only): target tokens with BTok KV as prefix}
        \State \(s^{(2)} \gets x^{(t)}\)
        \State \(h^{(2)}_0 \gets \mathrm{Embed}(s^{(2)})\)
        \For{\( \ell = 1 \) \textbf{to} \( L \)}
            \State
            \(\big(h^{(2)}_\ell, \_\big)
                \gets \mathrm{DecoderLayer}_\ell\big(
                    h^{(2)}_{\ell-1};
                    \text{past\_kv} = \text{KV}^{B}_\ell,
                    \text{mask} = \text{causal},
                    \text{use\_cache} = \texttt{False}
                \big)\)
            \State \textbf{// Each target token attends to all BToks (via past\_kv)}
            \State \textbf{// and previous target tokens (via causal mask)}
        \EndFor
        \State Compute token-level logits on \(h^{(2)}_L\) over textual target tokens
        \State \(\mathcal{L}_{\mathrm{ntp}} \gets
            \mathrm{NTP\_Loss}\big(h^{(2)}_L, x^{(t)}_{\text{text}}\big)\)
        \State \textbf{return} \(e(x), \mathcal{L}_{\mathrm{ntp}}\)
    \end{algorithmic}
\end{algorithm}

This procedure is equivalent to a single forward with the
Condensation Mask defined in \cref{sec:gic} (\cref{eq:mask}): Pass~1 realizes the query$\to$BTok
visibility, and Pass~2 realizes the BTok$\to$target and target$\to$target
visibility without ever exposing query tokens as KV during the target-side
pass.

\paragraph{Inference.}
At inference time, Pass~2 and the NTP loss are discarded entirely.
The model executes only Pass~1---a single standard causal forward
pass over the input tokens and BToks---and returns the mean-pooled
embedding $e(x)$.
This makes BToks inference identical to conventional last-token
pooling except for $K$ additional tokens, incurring negligible
overhead (see \cref{tab:efficiency}).

\section{Sensitivity Studies}
\label{sec:sensitivity}

Beyond the main configuration reported in \cref{sec:main_results}, we further study
the sensitivity of BToks to two key hyper-parameters: (i) the LoRA
rank used for parameter-efficient fine-tuning, and (ii) the scheduling
of the NTP loss weight $\lambda(t)$ in the joint objective. In all
experiments, we keep the backbone, data mixture, batch size, and
training schedule fixed and vary a single factor at a time.
All scores are macro-averaged following the MMEB-V2 protocol.

\paragraph{LoRA rank.}
\Cref{tab:lora_rank_sensitivity} reports the performance of BToks
on MMEB-V2 when varying the LoRA rank while keeping all other
hyper-parameters (including the default two-stage $\lambda(t)$
schedule) unchanged. Using a very low rank ($r=8$) leads to a notable
drop in the overall score (53.9) and especially hurts Video and VisDoc
performance. Increasing the rank to $r=16$ yields substantial gains
across all modalities and achieves the best overall and per-modality
scores. Further increasing the rank to $r=32$ produces results close
to, but consistently slightly below, the $r=16$ setting. These trends
suggest that BToks does not rely on very large adapters and that our
default rank offers a good trade-off between accuracy and parameter
efficiency.

\begin{table}[t]
  \centering
  \small
  \caption{Sensitivity of BToks to the LoRA rank on MMEB-V2.
  The default $r{=}16$ achieves the best overall and per-modality scores.}
  \label{tab:lora_rank_sensitivity}
\setlength{\tabcolsep}{8pt}
  \begin{tabular}{lcccc}
    \toprule
    LoRA rank & Overall & Image & Video & VisDoc \\
    \midrule
    $r = 8$  &  53.9 & 62.2  &  34.5 &  56.1 \\
    $r = 16$ (default) &  \textbf{59.0} &  \textbf{66.0} &  \textbf{39.9} &  \textbf{62.7} \\
    $r = 32$ &  58.5 &  65.6 &  39.3 &  62.4 \\
    \bottomrule
  \end{tabular}
\end{table}

\paragraph{NTP loss weight $\lambda(t)$.}
We also examine the effect of different choices of the NTP loss weight
schedule $\lambda(t)$ in the combined objective
$\mathcal{L}(t) = \mathcal{L}_{\mathrm{ctr}} + \lambda(t)\,
\mathcal{L}_{\mathrm{ntp}}$. The main configuration uses a two-stage
schedule: $\lambda(t)=0.1$ for the first $2{,}000$ steps and
$\lambda(t)=0$ for the remaining $3{,}000$ steps. As shown in
\cref{tab:lambda_sensitivity}, this schedule achieves the best
overall score (59.0) and the strongest performance on Video (39.9) and
VisDoc (62.7), indicating that a short early phase of
NTP under the Condensation Mask is sufficient to shape the bottleneck
interface.

All schedules with a non-zero $\lambda$ restricted to an early phase
(outside the constant $\lambda$ setting) outperform the purely
contrastive baseline ($\lambda(t)\equiv 0$) in terms of overall score.
For example, using $\lambda(t)=0.1$ for the first $1\mathrm{k}$ steps
already improves the overall score to 58.9 and yields the best Image
performance (66.6), but slightly lags behind the default schedule on
Video and VisDoc. In contrast, longer or stronger NTP phases (e.g.,
$\lambda(t)=0.2$ for $t\!<\!2\mathrm{k}$ or
$\lambda(t)=0.1$ for $t\!<\!3\mathrm{k}$/$4\mathrm{k}$) and a
constant $\lambda(t)\equiv 0.1$ lead to noticeable degradation,
suggesting that excessive NTP supervision can over-regularize the
learned representations. Overall, these results support
our design choice of a moderate, early NTP warm-up followed by purely
contrastive training over BToks.

\begin{table}[t]
  \centering
  \small
  \caption{Sensitivity of BToks to the NTP loss weight schedule
  $\lambda(t)$ on MMEB-V2. A short early NTP phase ($\lambda{=}0.1$, first $2\mathrm{k}$ steps) works best; longer or constant schedules degrade performance.}
  \label{tab:lambda_sensitivity}
\setlength{\tabcolsep}{4pt}
  \begin{tabular}{lcccc}
    \toprule
    $\lambda(t)$ schedule & Overall & Image & Video & VisDoc \\
    \midrule
    $\lambda(t) \equiv 0$ (contrastive-only)
      &  57.6 &  65.6 &  39.8 & 59.1 \\
    $0.1,\ t\!<\!1\mathrm{k}$;\ $0$ otherwise
      &  58.9 & \textbf{66.6} & 39.2 & 62.1 \\
    $0.05,\ t\!<\!2\mathrm{k}$;\ $0$ otherwise
      &  58.2 & 66.1 & 38.5 & 61.1 \\
    $0.1,\ t\!<\!2\mathrm{k}$;\ $0$ otherwise (default) &
      \textbf{59.0} &  66.0 &  \textbf{39.9} &  \textbf{62.7} \\
    $0.2,\ t\!<\!2\mathrm{k}$;\ $0$ otherwise
      &  58.3 & 66.3 & 38.9 & 60.7 \\
    $0.1,\ t\!<\!3\mathrm{k}$;\ $0$ otherwise
      & 57.5 & 65.2 & 38.1 & 60.6 \\
    $0.1,\ t\!<\!4\mathrm{k}$;\ $0$ otherwise
      & 56.9 & 63.3 & 38.3 & 61.1  \\
    $\lambda(t) \equiv 0.1$ (constant)
      & 56.3 & 64.0 & 37.3 & 59.0 \\
    \bottomrule
  \end{tabular}
\end{table}

\section{Per-dataset Results}
\label{sec:per_dataset}

\Cref{tab:per_dataset_image,tab:per_dataset_video,tab:per_dataset_visdoc}
show the full per-dataset breakdown of BToks and the VLM2Vec-V2
baseline on all 78 MMEB-V2 evaluation datasets, grouped by modality.
Each dataset is evaluated with its canonical metric (e.g., accuracy,
Hit@1, or NDCG@5), and modality-level scores are macro-averaged across
datasets.
BToks improves performance on the large majority of datasets, with the most pronounced gains on semantically demanding tasks such as GQA (+11.8), NExTQA (+26.0), and SynthDocQA-Energy (+10.3). The few regressions (e.g., HatefulMemes $-$7.0, Breakfast $-$6.3) occur on datasets where surface-level visual cues dominate over compositional reasoning.

\begin{table}[t]
  \centering
  \caption{Per-dataset results on \textbf{Image} tasks from MMEB-V2.
  Meta-task averages are shown in \textbf{bold}.}
  \label{tab:per_dataset_image}
\scriptsize
\setlength{\tabcolsep}{4pt}
  \begin{tabular}{llrrr}
    \toprule
    Meta-task & Dataset & VLM2Vec-V2 & BToks & $\Delta$ \\
    \midrule
    \multirow{11}{*}{Classification}
    & \textbf{Average} & \textbf{64.5} & \textbf{64.3} & \textbf{$-$0.2} \\
    & VOC2007        & 87.0 & 85.7 & $-$1.3 \\
    & N24News        & 79.9 & 74.0 & $-$5.9 \\
    & SUN397         & 72.7 & 74.9 & +2.2 \\
    & ObjectNet      & 60.0 & 68.5 & +8.5 \\
    & Country211     & 25.0 & 25.2 & +0.2 \\
    & Place365       & 39.7 & 38.7 & $-$1.0 \\
    & ImageNet-1K    & 78.7 & 80.5 & +1.8 \\
    & HatefulMemes   & 69.5 & 62.5 & $-$7.0 \\
    & ImageNet-A     & 43.7 & 46.9 & +3.2 \\
    & ImageNet-R     & 88.6 & 85.6 & $-$3.0 \\
    \midrule
    \multirow{11}{*}{Image QA}
    & \textbf{Average} & \textbf{56.2} & \textbf{59.8} & \textbf{+3.6} \\
    & OK-VQA         & 58.2 & 61.8 & +3.6 \\
    & A-OKVQA        & 50.2 & 48.6 & $-$1.6 \\
    & DocVQA         & 91.0 & 91.9 & +0.9 \\
    & InfographicsVQA & 58.7 & 61.6 & +2.9 \\
    & ChartQA        & 48.5 & 51.2 & +2.7 \\
    & Visual7W       & 50.6 & 49.2 & $-$1.4 \\
    & ScienceQA      & 34.0 & 40.1 & +6.1 \\
    & GQA            & 52.1 & 63.9 & +11.8 \\
    & TextVQA        & 74.6 & 79.5 & +4.9 \\
    & VizWiz         & 44.5 & 49.8 & +5.3 \\
    \midrule
    \multirow{13}{*}{Retrieval}
    & \textbf{Average} & \textbf{67.2} & \textbf{68.8} & \textbf{+1.6} \\
    & VisDial        & 78.4 & 78.4 & 0.0 \\
    & CIRR           & 49.9 & 54.0 & +4.1 \\
    & VisualNews\_t2i & 71.5 & 72.7 & +1.2 \\
    & VisualNews\_i2t & 74.4 & 75.8 & +1.4 \\
    & MSCOCO\_t2i    & 72.2 & 73.9 & +1.7 \\
    & MSCOCO\_i2t    & 67.8 & 69.1 & +1.3 \\
    & NIGHTS         & 66.7 & 68.1 & +1.4 \\
    & WebQA          & 89.5 & 90.5 & +1.0 \\
    & FashionIQ      & 16.2 & 19.1 & +2.9 \\
    & Wiki-SS-NQ     & 68.3 & 70.5 & +2.2 \\
    & OVEN           & 63.8 & 67.8 & +4.0 \\
    & EDIS           & 87.1 & 85.4 & $-$1.7 \\
    \midrule
    \multirow{5}{*}{Grounding}
    & \textbf{Average} & \textbf{74.7} & \textbf{77.4} & \textbf{+2.7} \\
    & MSCOCO         & 66.2 & 66.4 & +0.2 \\
    & RefCOCO        & 83.0 & 87.2 & +4.2 \\
    & RefCOCO-Match  & 86.3 & 86.4 & +0.1 \\
    & Visual7W-Point & 63.4 & 69.4 & +6.0 \\
    \midrule
    \multicolumn{2}{l}{\textbf{Image Overall}} & \textbf{64.2} & \textbf{66.0} & \textbf{+1.8} \\
    \bottomrule
  \end{tabular}
\end{table}

\begin{table}[t]
  \centering
  \caption{Per-dataset results on \textbf{Video} tasks from MMEB-V2.
  Meta-task averages are shown in \textbf{bold}.}
  \label{tab:per_dataset_video}
\scriptsize
\setlength{\tabcolsep}{4pt}
  \begin{tabular}{llrrr}
    \toprule
    Meta-task & Dataset & VLM2Vec-V2 & BToks & $\Delta$ \\
    \midrule
    \multirow{6}{*}{Classification}
    & \textbf{Average} & \textbf{39.1} & \textbf{43.7} & \textbf{+4.6} \\
    & K700           & 33.8 & 43.1 & +9.3 \\
    & UCF101         & 59.8 & 69.3 & +9.5 \\
    & HMDB51         & 38.5 & 47.1 & +8.6 \\
    & SmthSmthV2     & 39.2 & 41.0 & +1.8 \\
    & Breakfast      & 24.3 & 18.0 & $-$6.3 \\
    \midrule
    \multirow{6}{*}{Video QA}
    & \textbf{Average} & \textbf{34.4} & \textbf{47.0} & \textbf{+12.6} \\
    & Video-MME      & 29.1 & 39.9 & +10.8 \\
    & MVBench        & 34.4 & 45.4 & +11.0 \\
    & NExTQA         & 21.9 & 47.9 & +26.0 \\
    & EgoSchema      & 34.8 & 37.0 & +2.2 \\
    & ActivityNetQA  & 51.6 & 64.8 & +13.2 \\
    \midrule
    \multirow{6}{*}{Retrieval}
    & \textbf{Average} & \textbf{28.2} & \textbf{33.0} & \textbf{+4.8} \\
    & MSR-VTT        & 28.7 & 33.8 & +5.1 \\
    & MSVD           & 49.6 & 56.0 & +6.4 \\
    & DiDeMo         & 30.1 & 33.0 & +2.9 \\
    & VATEX          & 23.1 & 27.6 & +4.5 \\
    & YouCook2       &  9.5 & 14.5 & +5.0 \\
    \midrule
    \multirow{4}{*}{Moment Ret.}
    & \textbf{Average} & \textbf{32.0} & \textbf{33.5} & \textbf{+1.5} \\
    & QVHighlight    & 40.0 & 42.2 & +2.2 \\
    & Charades-STA   & 16.1 & 18.0 & +1.9 \\
    & MomentSeeker   & 40.1 & 40.4 & +0.3 \\
    \midrule
    \multicolumn{2}{l}{\textbf{Video Overall}} & \textbf{33.6} & \textbf{39.9} & \textbf{+6.3} \\
    \bottomrule
  \end{tabular}
\end{table}

\begin{table}[t]
  \centering
  \caption{Per-dataset results on \textbf{Visual Document} tasks from MMEB-V2.
  Meta-task averages are shown in \textbf{bold}.}
  \label{tab:per_dataset_visdoc}
\scriptsize
\setlength{\tabcolsep}{4pt}
  \begin{tabular}{llrrr}
    \toprule
    Meta-task & Dataset & VLM2Vec-V2 & BToks & $\Delta$ \\
    \midrule
    \multirow{11}{*}{ViDoRe-V1}
    & \textbf{Average} & \textbf{66.0} & \textbf{71.1} & \textbf{+5.1} \\
    & ArxivQA        & 69.5 & 76.5 & +7.0 \\
    & DocVQA         & 30.9 & 37.2 & +6.3 \\
    & InfoVQA        & 78.4 & 80.9 & +2.5 \\
    & TabFQuAD       & 74.8 & 80.4 & +5.6 \\
    & TATDQA         & 35.6 & 42.0 & +6.4 \\
    & ShiftProject   & 56.1 & 62.3 & +6.2 \\
    & SynthDocQA-AI  & 80.4 & 79.9 & $-$0.5 \\
    & SynthDocQA-Energy & 75.0 & 85.3 & +10.3 \\
    & SynthDocQA-Gov & 75.7 & 81.0 & +5.3 \\
    & SynthDocQA-Health & 83.2 & 85.3 & +2.1 \\
    \midrule
    \multirow{5}{*}{ViDoRe-V2}
    & \textbf{Average} & \textbf{37.1} & \textbf{38.6} & \textbf{+1.5} \\
    & ESG-Reports    & 41.7 & 40.0 & $-$1.7 \\
    & BioMed-Lect    & 34.3 & 39.1 & +4.8 \\
    & Econ-Reports   & 31.2 & 39.7 & +8.5 \\
    & ESG-V2         & 41.1 & 35.7 & $-$5.4 \\
    \midrule
    \multirow{7}{*}{VisRAG}
    & \textbf{Average} & \textbf{77.6} & \textbf{81.3} & \textbf{+3.7} \\
    & ArxivQA        & 73.0 & 76.9 & +3.9 \\
    & ChartQA        & 77.8 & 86.2 & +8.4 \\
    & MP-DocVQA      & 77.0 & 78.6 & +1.6 \\
    & SlideVQA       & 89.8 & 91.8 & +2.0 \\
    & InfoVQA        & 86.4 & 88.7 & +2.3 \\
    & PlotQA         & 61.5 & 65.6 & +4.1 \\
    \midrule
    \multirow{5}{*}{VisDoc-OOD}
    & \textbf{Average} & \textbf{32.4} & \textbf{38.1} & \textbf{+5.7} \\
    & ViDoSeek-page  & 17.1 & 21.7 & +4.6 \\
    & ViDoSeek-doc   & 73.4 & 78.3 & +4.9 \\
    & MMLongBench-page &  8.0 & 11.0 & +3.0 \\
    & MMLongBench-doc & 31.1 & 41.4 & +10.3 \\
    \midrule
    \multicolumn{2}{l}{\textbf{VisDoc Overall}} & \textbf{58.5} & \textbf{62.7} & \textbf{+4.2} \\
    \bottomrule
  \end{tabular}
\end{table}

\end{document}